\documentclass{article}

    \usepackage[preprint]{neurips_2019}

\usepackage[utf8]{inputenc} % allow utf-8 input
\usepackage[T1]{fontenc}    % use 8-bit T1 fonts
\usepackage{hyperref}       % hyperlinks
\usepackage{url}            % simple URL typesetting
\usepackage{booktabs}       % professional-quality tables
\usepackage{amsfonts}       % blackboard math symbols
\usepackage{nicefrac}       % compact symbols for 1/2, etc.
\usepackage{microtype}      % microtypography
\usepackage[pdftex]{graphicx}
\usepackage{todonotes}
\usepackage{ulem}
\usepackage{soul}

\usepackage{amsmath}
\usepackage{natbib}
\usepackage{subcaption}
\usepackage[table]{colortbl}
\usepackage{caption} 
\usepackage[noend]{algpseudocode}
\usepackage{algorithm}

\DeclareMathOperator*{\argmax}{arg\,max}
\DeclareMathOperator*{\argmin}{arg\,min}



\title{Developmentally motivated emergence of compositional communication via template transfer}

\author{
  Tomasz Korbak \\
  Institute of Philosophy and Sociology,\\
  Polish Academy of Sciences \vspace{3pt}\\
   Human Interactivity and Language Lab, \\
Faculty of Psychology, \\
University of Warsaw, Poland \vspace{3pt}\\
  \texttt{tomasz.korbak@gmail.com} \\
   \And
   Julian Zubek \\
   Human Interactivity and Language Lab, \\
Faculty of Psychology, \\
University of Warsaw, Poland \vspace{3pt}\\
   \texttt{j.zubek@uw.edu.pl} \\
   \And
   Łukasz Kuciński \\
   Institute of Mathematics, \\
   Polish Academy of Sciences \vspace{3pt}\\
   \texttt{lukasz.kucinski@impan.pl} \\
   \And
   Piotr Miłoś \\
   Institute of Mathematics, \\
   Polish Academy of Sciences, \vspace{3pt}\\
   deepsense.ai \vspace{3pt}\\
   \texttt{pmilos@mimuw.edu.pl} \\
   \And
   Joanna Rączaszek-Leonardi \\
   Human Interactivity and Language Lab, \\
Faculty of Psychology, \\
University of Warsaw, Poland \vspace{3pt}\\
   \texttt{raczasze@psych.uw.edu.pl} \\
}

\begin{document}

\maketitle
\begin{abstract}
  This paper explores a novel approach to achieving emergent compositional communication in multi-agent systems. We propose a training regime implementing \textit{template transfer}, the idea of carrying over learned biases across contexts. In our method, a sender--receiver pair is first trained with disentangled loss functions and then the receiver is \textit{transferred} to train a new sender with a standard loss. Unlike other methods (e.g. the obverter algorithm), our approach does not require imposing inductive biases on the architecture of the agents. We experimentally show the emergence of compositional communication using topographical similarity, zero-shot generalization and context independence as evaluation metrics. The presented approach is connected to an important line of work in semiotics and developmental psycholinguistics: it supports a conjecture that compositional communication is scaffolded on simpler communication protocols.
\end{abstract}

\vspace{-5pt} 
\section{Introduction}
\vspace{-5pt} 

Language-like communication protocols can arise in environments that require agents to share information and coordinate behavior \citep{foerster_learning_2016,lazaridou_multi-agent_2016,jaques_social_2018}. A crucial feature of human languages and some animal communication systems is {\it compositionality}: there are complex signals constructed through the combination of signals. Furthermore, compositionality is essential for general intelligence because it facilitates generalization ---adaptability to novel situations---and productivity---an infinite number of meanings can be created using a finite set of primitives \citep{lake_building_2016}. 

The contribution of this paper is its demonstration that communication protocols exhibiting compositionality can emerge via adaptation of pre-existing, simpler non-compositional protocols to a new environment. This procedure is an instance of {\it template transfer} \citep{barrett_self-assembling_2017}. 

We are motivated in this work by semiotic theoretical framework and empirical research in language development. \cite{deacon_symbolic_1998}, after \cite{peirce_collected_1998} presented a semiotic framework detailing how (i) iconic and (ii) indexical communication protocols precede (iii) complex compositional communication protocols. Similarly, language development research demonstrates that children learn to speak compositionally in a structured social environment designed for teaching progressively more complex utterances through simple language games \citep{stern_goal_1974,bruner_childs_1983,nomikou_verbs_2017, raczaszek-leonardi_language_2018}.

Our model implements the idea of template transfer by sharing agents across games of varying complexity. We decompose learning compositional communication into three phases: (i) learning a visual classifier, (ii) learning non-compositional communication protocols, and (iii) learning a compositional communication protocol. This decomposition closely follows distinctions established in semiotics and is more plausible in the light of human language development than other approaches. Crucially, the biases learned in simple games in phase (ii) are sufficient to incentivize a compositional communication protocol to emerge in phase (iii). We compare the template transfer approach with other method of achieving compositionality---the obverter algorithm \citep{batali_computational_1998,choi_compositional_2018}---on three different metrics: zero-shot generalization, context independence and topographical similarity. The results demonstrate that the ability to communicate compositionality can emerge in a model less cognitively demanding than the obverter approach. The key idea---that complex communication protocols are easier to learn when bootstrapped on pre-existing simpler protocols---may generalize to other problems in multi-agent communication.
\vspace{-5pt} 
\section{Related work}
\vspace{-5pt} 
Recent work on emergent communication in artificial intelligence shows that compositionality requires strong inductive biases to be imposed on communicating agents \citep{kottur_natural_2017}. One recurring idea is placing pressure on agents to use symbols consistently across varying contexts. To that end, \cite{kottur_natural_2017} and \cite{das_learning_2017} reset the memory of an agent between producing or receiving subsequent symbols, which helps to obtain a consistent symbol grounding. A more psychologically plausible approach is explored by \cite{choi_compositional_2018} and \cite{bogin_emergence_2018}, who take inspiration from the obverter algorithm \citep{oliphant_learning_1997,batali_computational_1998}.

The obverter (from the Latin {\it obverto}, to turn towards) algorithm \citep{batali_computational_1998,oliphant_learning_1997} is based on the assumption that an agent can use its own responses to messages to predict other agent's responses, and thus can iteratively compose its messages to maximize the probability of the desired response. A limitation of the obverter is that it makes strong assumptions about the agents and task: to be able to use themselves as models of others, the agents must share an identical architecture and the task must be symmetric (the agents must be able to exchange their roles). This excludes games with functional specialization of agents. 
% Another problem is computational complexity of the decoding procedure. Even assuming greedy decoding, producing a message requires $\mathcal{O}(vT)$ queries to the model of the receiver (where $v$ is vocabulary size and $T$ is maximum message length).

A different family of approaches, more similar in spirit to the approach described in this work, includes population-based training that incentivizes the creation of communication protocols that are easy to teach to new agents \citep{brighton_compositional_2002,li_ease--teaching_2019} and gradually increasing task complexity that incentivizes reusing existing patterns of communication \citep{de_beule_emergence_2006}. Contributing to this line of thinking, we show how the history of acquiring simpler communication protocols may lead to more complex types of communication.

\vspace{-5pt} 
\section{Method}
\vspace{-5pt} 
In this section we present two Lewis signaling games \citep{lewis_convention:_2011,skyrms_signals:_2010}. The first is a standard \textit{object naming game}. Based on it, we propose a new game that incorporates the idea of template transfer \citep{barrett_self-assembling_2017}.
\vspace{-2pt} 
\paragraph{Object naming game} Two agents, a sender and a receiver, learn to communicate about colored geometric objects. The sender observes an object (an RGB image) and sends a message (a sequence of discrete symbols) to the receiver; the receiver must correctly indicate both the color and the shape of the object. Formally, the game is stated as maximization of the following log likelihood:
\vspace{-2pt} 
\begin{equation*}
\label{loss}
    \mathcal{L}(\theta, \psi) := \mathbb{E}_{x} \mathbb{E}_{m \sim s_{\theta}(\cdot \vert x)} [-\log r_{\psi}(x_c, x_s \vert m)],
\end{equation*}
\vspace{-2pt} 
where $s_{\theta}$ is the policy of the sender (i.e. $s_{\theta}(m|x)$ is the probability of sending message $m$ when observing image $x$), $r_{\psi}$ is the policy of the receiver, $x$ is the representation of the object and $x_c$ and $x_s$ are its color and shape, respectively. Parameters $\theta$ and $\psi$ are learnable parameters of the polices. For more details, see Algorithm \ref{algo-baseline} in the Appendix.

\paragraph{Template transfer game} This game consists of two phases. In the first, two senders communicate with one receiver in two sub-games. The sub-games are disentangled in the sense that their tasks are to correctly indicate one aspect of the object (color or shape), as formalized by the following loss functions:
\vspace{-3pt}
\begin{equation*}
\label{pretraining-loss}
    \mathcal{L}_1(\theta_1, \psi) := \mathbb{E}_{x} \mathbb{E}_{m \sim s_{\theta_1}(\cdot \vert x)} [-\log r_{\psi}(x_c \vert m)],\quad \mathcal{L}_2(\theta_2, \psi) := \mathbb{E}_{x} \mathbb{E}_{m \sim s_{\theta_2}(\cdot \vert x)} [-\log r_\psi(x_s \vert m)],
\end{equation*}
\vspace{-2pt}
where $r_\theta(x_c|m)$ is the marginalization of $r_\theta(x_c, x_s|m)$, viz. $r_\theta(x_c|m):=\sum_{x_s} r_\theta(x_c, x_s|m)$. Analogously, one can define $r_\theta(x_c|m)$.

These two losses are optimized simultaneously (crucially with the shared parameters $\psi$ of the receiver) until a desired level of accuracy is met. Then, the second phase follows, in which the receiver is passed (via template transfer) to the object naming game (as described in the previous paragraph) with a new sender. See Figure \ref{fig:procedure} and Algorithm \ref{algo-template transfer} in the Appendix for more details.

The communication protocol acquired in the first phase serves as a training bias in the second phase. Informally, the new sender learns to emulate messages sent by the two specialized senders of the previous phase. Our experiments (Section \ref{sec:experiments_and_results}) indicate that two-phase learning is a sufficient incentive for compositionality to emerge. 
\begin{figure}[h]
\centering
\includegraphics[width=0.5\linewidth]{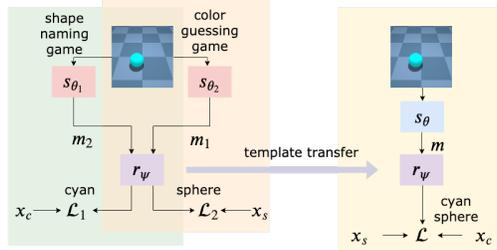}
\caption{Template transfer consists of pre-training the receiver $r_\psi$ on two games with disentangled losses $\mathcal{L}_1$ and $\mathcal{L}_2$ and transferring $r_\psi$ to a new object naming game.}
\label{fig:procedure}
\end{figure}
\vspace{-15pt} 
\paragraph{Agents} Both the sender and the receiver are implemented as recurrent neural networks. The sender is equipped with a pre-trained convolutional neural network to process visual input. After observing the object, the sender generates a sequence of $T$ discrete messages sampled from a closed vocabulary of 10 symbols. During the object naming game, $T = 2$, while during the template transfer game, $T = 1$. To prevent distribution shift with respect to message length between games, a random uniformly sampled symbol is prepended to $s_1$'s messages and appended to $s_2$'s messages. Due to the discrete nature of the communication channel, Gumbel--Softmax relaxation \citep{maddison_concrete_2016,jang_categorical_2016} is used to backpropagate through $s_\theta(m \vert x)$. For details concerning the architecture and hyperparameters, please see the Appendix. \vspace{-2pt} 
\vspace{-3pt} 
\section{Experiments and results}\label{sec:experiments_and_results}
\vspace{-5pt} 
\paragraph{Measuring compositionality}
We utilize three metrics of compositionality of a communication protocol: \textit{zero-shot generalization accuracy}, \textit{context independence} and \textit{topographical similarity}. During evaluation we use the deterministic sender given by $s(x) := \argmax_m s_\theta(m|x)$, where $x$ is an object.

We quantify zero-shot generalization by measuring the accuracy of the agents on a test set, containing pairs of shapes and colors not present in the training set. 

Context independence was introduced by \cite{bogin_emergence_2018} as a measure of alignment between the symbols in an agent’s messages and the concepts transmitted. We denote the set of symbols used to compose messages by $V$ and by $K$ the set of concepts, which in our case is the union of available colors and shapes. Given sender $s$, and assuming a uniform distribution of objects, we define $p(v|k)$ as the probability that symbol $v\in V$ appears when the sender observes an object with property $k\in K$. We define $p(k|v)$ in the same manner. Further, let $v^k := \argmax_v p(k \vert v)$. The context independence metric is defined as $\mathbb{E} ( p(v^k \vert k) \cdot p(k \vert v^k))$; the expectation is taken with respect to the uniform distribution on $K$. Context independence is sometimes considered restrictive, as it may be low in the presence of synonyms \citep{lowe_pitfalls_2019}.

Finally, we introduce topographical similarity \citep{brighton_understanding_2006,lazaridou_emergence_2018}, also known as \textit{representational similarity} \citep{kriegeskorte_representational_2008,bouchacourt_how_2018}. We denote the random variable $L_t:=L((x^1_c, x^1_s), (x^2_c, x^2_s))$, where $L$ is the \cite{levenshtein_binary_1966} distance and $x^1$ and $x^2$ are independently sampled objects with the subscripts denoting their shapes and colors. Note that in our case $L_t\in \{0, 1, 2\}$. Let $L_m:=L(s(x^1), s(x^2))$ be the distance between messages sent by the sender after observing $x^1$ and $x^1$. Topographical similarity is the the Spearman $\rho$ correlation of $L_t$ and $L_m$.

The metrics described above can be regarded as measures of compositionality. Indeed, high zero-shot generalization indicates that the agents correctly map the implicit compositional structure of inputs to explicate one of the outputs. The other two metrics focus directly on the transmitted messages, comparing them to the ground truth, fully disentangled (color, shape) representation.  
\vspace{-4pt} 
\paragraph{Obverter baseline} In the obverter algorithm, two agents exchange the roles of the sender and the receiver. If an agent is the receiver, it behaves as in the object naming game. If an agent is the sender, it sends message that would have produced the most accurate prediction of color and shape, if it had received such a message as a receiver (i.e. instead of the greedy decoding used in the original implementation of \cite{batali_computational_1998}, we simply choose the message maximizing accuracy). Accuracy is evaluated against the predictions of the visual classifier. For details, consult Algorithm \ref{algo-obverter} in the Appendix.
\vspace{-4pt} 
\paragraph{The effect of template transfer on compositionality}

We compared our approach with several baselines (random, the same architecture without pre-training games, and our implementation of the obverter approach) on games with five shapes and five colors\footnote{The code accompanying this paper is released on GitHub under \url{https://github.com/tomekkorbak/compositional-communication-via-template-transfer}.}. Topographical similarity and context independence were computed on the full dataset (train and test); the results are presented in Table \ref{main-results}. Template transfer clearly leads to highly compositional communication protocols. While all methods struggled to generalize to unseen objects, template transfer was the most successful. For examples of communication protocols representative of the experiments conducted, see Table \ref{protocols-examples}.
% \vspace{-5}
\begin{table}[h!]
% \vspace{-10pt}
  \caption{The effect of template transfer on compositionality. The metrics are train and test set accuracies (the rate of correctly predicted both $x_c$ and $x_s$); average over the individual accuracies for $x_c$ and $x_s$; and context independence (CI) and topographical similarity (Topo). The models are random baseline (untrained agents); baseline architecture (without template transfer); template transfer (TT); and obverter algorithm.  All reported metrics are averaged over ten random seeds and standard deviations are reported in brackets.
  \vspace{-5pt}}
  \label{main-results}
  \centering
  \begin{tabular}{llllll}
    \toprule
    & \multicolumn{3}{c}{Accuracy} & \\
    \cmidrule(r){2-4}
    Model & Train (both) & Test (both) & Test (avg) & CI & Topo \\
    \midrule
    Random  & 0.04 & 0.04 & 0.2  & 0.04 ($\pm$ 0.01) & 0.13 ($\pm$ 0.03) \\
    Baseline  & 0.99 ($\pm$ 0.01) & 0.02 ($\pm$ 0.05) & 0.47 ($\pm$ 0.09) &  0.08 ($\pm$ 0.01)& 0.30 ($\pm$ 0.05) \\
    Obverter  & 0.99 ($\pm$ 0) & 0.24 ($\pm$ 0.23) & 0.51 ($\pm$ 0.19) & 0.12 ($\pm$ 0.02) & 0.55 ($\pm$ 0.13) \\
    TT (ours) & 1 ($\pm$ 0) & 0.48 ($\pm$ 0.10) & 0.74 ($\pm$ 0.06) & 0.18 ($\pm$ 0.01) & 0.85 ($\pm$ 0.03) \\
    \bottomrule
  \end{tabular}
\end{table}
\addtolength{\tabcolsep}{-4pt}   
\begin{table}[htb]
\centering
% \vspace{-16pt}
\caption{Two example communication protocols, one that emerged via the baseline architecture (\ref{table-noncomp}), and one via template transfer (\ref{table-noncomp}). Gray cells indicate objects not seen during training. In  (\ref{table-comp}), symbols exhibit clear association with colors and shapes, e.g. symbol 8 is consistently associated with the color magenta (when on first position) and boxes (when on second position).}
\vspace{-10pt}
\begin{tabular}{ll}
% \vspace{-15pt}
\begin{subtable}{.5\textwidth}\centering
\subcaption{A non-compositional communication protocol \\ (topographical similarity 0.25)}
\label{table-noncomp}
% \vspace{-5pt}
\begin{tabular}{rccccc}
  & box & sphere & cylinder & torus & ellipsoid    \\
 blue & \cellcolor{gray!25}1 0 & 4 5       & 1 0  & 4 5  & 5 0       \\
 cyan & 9 0       & \cellcolor{gray!25} 4 0 & 3 0   & 4 0      & 7 0      \\
 gray & 3 5       & 6 5       & \cellcolor{gray!25}3 2 & 6 5  & 5 3  \\
 green & 0 0       & 7 6   & 3 0       & \cellcolor{gray!25}6 0     & 7 6      \\
 magenta & 1 5      & 5 5       & 1 2  & 1 5  & \cellcolor{gray!25}5 2  \\
\end{tabular}
\end{subtable}
\begin{subtable}{.5\textwidth}\centering
\subcaption{A highly compositional communication protocol \\ (topographical similarity 0.85)}
\label{table-comp}
% \vspace{-5pt}
\begin{tabular}{rccccc}
  & box & sphere & cylinder & torus & ellipsoid    \\
  blue & \cellcolor{gray!25}1 8& 1 9  & 1 5  & 1 6  & 1 4  \\
 cyan & 4 8            & \cellcolor{gray!25}4 9 & 4 5  & 4 6  & 4 4  \\
 gray & 6 8            & 6 9  & \cellcolor{gray!25}6 5 & 6 6  & 6 9  \\
 green & 9 8            & 9 9  & 9 5  & \cellcolor{gray!25}9 6 & 9 4  \\
 magenta & 8 8            & 8 9  & 8 5  & 8 8  & \cellcolor{gray!25}8 4 \\
\end{tabular}
\end{subtable}
\end{tabular}
\label{protocols-examples}
\end{table}
\vspace{-3pt} 
\section{Conclusions}
\vspace{-3pt} 
Template transfer — transferring skills from simpler to more complex Lewis signaling games — can be used to model a variety of semiotic, social and cognitive phenomena \citep{barrett_self-assembling_2017,barrett_self-assembling_2019,barrett_hierarchical_2018}. In this paper we demonstrated that compositional communication can emerge via template transfer, presenting a model-free approach more general than the obverter algorithm. The one assumption that we make is that the loss function can be decomposed into two disentangled loss functions, as in the case of decomposing $\mathcal{L}$ into $\mathcal{L}_1$ and $\mathcal{L}_2$. (Note that there is no need for inputs to be disentangled.)  This shows that biases necessary for compositional communication can be learned rather than imposed by design.

\subsubsection*{Acknowledgments}

Tomasz Korbak was funded by Ministry of Science and Higher Education (Poland) grant DI2015010945. Joanna Rączaszek-Leonardi and Julian Zubek were funded by National Science Centre (Poland) grant OPUS 2018/29/B/HS1/00884. Piotr Miłoś was funded by National Science Centre (Poland) grant UMO-2017/26/E/ST6/0062. The computations were run on the PLGRID infrastructure within the Prometheus cluster.

\bibliography{references}

\begin{thebibliography}{}

\bibitem[\protect\astroncite{Barrett et~al.}{2018}]{barrett_hierarchical_2018}
Barrett, J., Skyrms, B., and Cochran, C. (2018).
\newblock Hierarchical {Models} for the {Evolution} of {Compositional}
  {Language}.

\bibitem[\protect\astroncite{Barrett and
  Skyrms}{2017}]{barrett_self-assembling_2017}
Barrett, J.~A. and Skyrms, B. (2017).
\newblock Self-assembling {Games}.
\newblock {\em The British Journal for the Philosophy of Science},
  68(2):329--353.

\bibitem[\protect\astroncite{Barrett
  et~al.}{2019}]{barrett_self-assembling_2019}
Barrett, J.~A., Skyrms, B., and Mohseni, A. (2019).
\newblock Self-{Assembling} {Networks}.
\newblock {\em The British Journal for the Philosophy of Science},
  70(1):301--325.

\bibitem[\protect\astroncite{Batali}{1998}]{batali_computational_1998}
Batali, J. (1998).
\newblock Computational simulations of the emergence of grammar.
\newblock {\em Approach to the Evolution of Language}, pages 405--426.

\bibitem[\protect\astroncite{Bogin et~al.}{2018}]{bogin_emergence_2018}
Bogin, B., Geva, M., and Berant, J. (2018).
\newblock Emergence of {Communication} in an {Interactive} {World} with
  {Consistent} {Speakers}.
\newblock {\em arXiv:1809.00549 [cs]}.
\newblock arXiv: 1809.00549.

\bibitem[\protect\astroncite{Bouchacourt and
  Baroni}{2018}]{bouchacourt_how_2018}
Bouchacourt, D. and Baroni, M. (2018).
\newblock How agents see things: {On} visual representations in an emergent
  language game.
\newblock {\em arXiv:1808.10696 [cs]}.
\newblock arXiv: 1808.10696.

\bibitem[\protect\astroncite{Brighton}{2002}]{brighton_compositional_2002}
Brighton, H. (2002).
\newblock Compositional {Syntax} {From} {Cultural} {Transmission}.
\newblock {\em Artificial Life}, 8(1):25--54.

\bibitem[\protect\astroncite{Brighton and
  Kirby}{2006}]{brighton_understanding_2006}
Brighton, H. and Kirby, S. (2006).
\newblock Understanding {Linguistic} {Evolution} by {Visualizing} the
  {Emergence} of {Topographic} {Mappings}.
\newblock {\em Artificial Life}, 12(2):229--242.

\bibitem[\protect\astroncite{Bruner}{1983}]{bruner_childs_1983}
Bruner, J.~S. (1983).
\newblock {\em Child's talk: learning to use language}.
\newblock W.W. Norton, New York, 1st ed edition.

\bibitem[\protect\astroncite{Choi et~al.}{2018}]{choi_compositional_2018}
Choi, E., Lazaridou, A., and de~Freitas, N. (2018).
\newblock Compositional {Obverter} {Communication} {Learning} {From} {Raw}
  {Visual} {Input}.
\newblock {\em ICLR 2018}.
\newblock arXiv: 1804.02341.

\bibitem[\protect\astroncite{Das et~al.}{2017}]{das_learning_2017}
Das, A., Kottur, S., Moura, J. M.~F., Lee, S., and Batra, D. (2017).
\newblock Learning {Cooperative} {Visual} {Dialog} {Agents} with {Deep}
  {Reinforcement} {Learning}.
\newblock {\em 2017 IEEE International Conference on Computer Vision (ICCV),
  Venice, 2017}.
\newblock arXiv: 1703.06585.

\bibitem[\protect\astroncite{De~Beule and
  Bergen}{2006}]{de_beule_emergence_2006}
De~Beule, J. and Bergen, B.~K. (2006).
\newblock On the emergence of compositionality.
\newblock In {\em The {Evolution} of {Language}}, pages 35--42, Rome, Italy.
  World scientific.

\bibitem[\protect\astroncite{Deacon}{1998}]{deacon_symbolic_1998}
Deacon, T.~W. (1998).
\newblock {\em The symbolic species: the co-evolution of language and the
  brain}.
\newblock Norton, New York, NY, norton paperback edition.
\newblock OCLC: 254499872.

\bibitem[\protect\astroncite{Foerster et~al.}{2016}]{foerster_learning_2016}
Foerster, J.~N., Assael, Y.~M., de~Freitas, N., and Whiteson, S. (2016).
\newblock Learning to {Communicate} with {Deep} {Multi}-{Agent} {Reinforcement}
  {Learning}.
\newblock {\em NIPS'16 Proceedings of the 30th International Conference on
  Neural Information Processing Systems}.
\newblock arXiv: 1605.06676.

\bibitem[\protect\astroncite{Jang et~al.}{2016}]{jang_categorical_2016}
Jang, E., Gu, S., and Poole, B. (2016).
\newblock Categorical {Reparameterization} with {Gumbel}-{Softmax}.
\newblock {\em arXiv:1611.01144 [cs, stat]}.
\newblock arXiv: 1611.01144.

\bibitem[\protect\astroncite{Jaques et~al.}{2018}]{jaques_social_2018}
Jaques, N., Lazaridou, A., Hughes, E., Gulcehre, C., Ortega, P.~A., Strouse,
  D.~J., Leibo, J.~Z., and de~Freitas, N. (2018).
\newblock Social {Influence} as {Intrinsic} {Motivation} for {Multi}-{Agent}
  {Deep} {Reinforcement} {Learning}.
\newblock {\em arXiv:1810.08647 [cs, stat]}.
\newblock arXiv: 1810.08647.

\bibitem[\protect\astroncite{Kharitonov et~al.}{2019}]{kharitonov_egg:_2019}
Kharitonov, E., Chaabouni, R., Bouchacourt, D., and Baroni, M. (2019).
\newblock {EGG}: a toolkit for research on {Emergence} of {lanGuage} in
  {Games}.
\newblock {\em arXiv:1907.00852 [cs]}.
\newblock arXiv: 1907.00852.

\bibitem[\protect\astroncite{Kingma and Ba}{2014}]{kingma_adam:_2014}
Kingma, D.~P. and Ba, J. (2014).
\newblock Adam: {A} {Method} for {Stochastic} {Optimization}.
\newblock {\em arXiv:1412.6980 [cs]}.
\newblock arXiv: 1412.6980.

\bibitem[\protect\astroncite{Kottur et~al.}{2017}]{kottur_natural_2017}
Kottur, S., Moura, J. M.~F., Lee, S., and Batra, D. (2017).
\newblock Natural {Language} {Does} {Not} {Emerge} '{Naturally}' in
  {Multi}-{Agent} {Dialog}.
\newblock {\em arXiv:1706.08502 [cs]}.
\newblock arXiv: 1706.08502.

\bibitem[\protect\astroncite{Kriegeskorte}{2008}]{kriegeskorte_representational_2008}
Kriegeskorte, N. (2008).
\newblock Representational similarity analysis – connecting the branches of
  systems neuroscience.
\newblock {\em Frontiers in Systems Neuroscience}.

\bibitem[\protect\astroncite{Lake et~al.}{2016}]{lake_building_2016}
Lake, B.~M., Ullman, T.~D., Tenenbaum, J.~B., and Gershman, S.~J. (2016).
\newblock Building {Machines} {That} {Learn} and {Think} {Like} {People}.
\newblock {\em arXiv:1604.00289 [cs, stat]}.
\newblock arXiv: 1604.00289.

\bibitem[\protect\astroncite{Lazaridou et~al.}{2018}]{lazaridou_emergence_2018}
Lazaridou, A., Hermann, K.~M., Tuyls, K., and Clark, S. (2018).
\newblock Emergence of {Linguistic} {Communication} from {Referential} {Games}
  with {Symbolic} and {Pixel} {Input}.
\newblock {\em arXiv:1804.03984 [cs]}.
\newblock arXiv: 1804.03984.

\bibitem[\protect\astroncite{Lazaridou
  et~al.}{2016}]{lazaridou_multi-agent_2016}
Lazaridou, A., Peysakhovich, A., and Baroni, M. (2016).
\newblock Multi-{Agent} {Cooperation} and the {Emergence} of ({Natural})
  {Language}.
\newblock {\em arXiv:1612.07182 [cs]}.
\newblock arXiv: 1612.07182.

\bibitem[\protect\astroncite{Levenshtein}{1966}]{levenshtein_binary_1966}
Levenshtein, V.~I. (1966).
\newblock Binary {Codes} {Capable} of {Correcting} {Deletions}, {Insertions}
  and {Reversals}.
\newblock {\em Soviet Physics Doklady}, 10:707.

\bibitem[\protect\astroncite{Lewis}{2011}]{lewis_convention:_2011}
Lewis, D.~K. (2011).
\newblock {\em Convention: a philosophical study}.
\newblock Blackwell, Oxford, nachdr. edition.
\newblock OCLC: 837747718.

\bibitem[\protect\astroncite{Li and Bowling}{2019}]{li_ease--teaching_2019}
Li, F. and Bowling, M. (2019).
\newblock Ease-of-{Teaching} and {Language} {Structure} from {Emergent}
  {Communication}.
\newblock {\em arXiv:1906.02403 [cs]}.
\newblock arXiv: 1906.02403.

\bibitem[\protect\astroncite{Lowe et~al.}{2019}]{lowe_pitfalls_2019}
Lowe, R., Foerster, J., Boureau, Y.-L., Pineau, J., and Dauphin, Y. (2019).
\newblock On the {Pitfalls} of {Measuring} {Emergent} {Communication}.
\newblock {\em arXiv:1903.05168 [cs, stat]}.
\newblock arXiv: 1903.05168.

\bibitem[\protect\astroncite{Maddison et~al.}{2016}]{maddison_concrete_2016}
Maddison, C.~J., Mnih, A., and Teh, Y.~W. (2016).
\newblock The {Concrete} {Distribution}: {A} {Continuous} {Relaxation} of
  {Discrete} {Random} {Variables}.
\newblock {\em arXiv:1611.00712 [cs, stat]}.
\newblock arXiv: 1611.00712.

\bibitem[\protect\astroncite{Nomikou et~al.}{2017}]{nomikou_verbs_2017}
Nomikou, I., Koke, M., and Rohlfing, K.~J. (2017).
\newblock Verbs in {Mothers}’ {Input} to {Six}-{Month}-{Olds}: {Synchrony}
  between {Presentation}, {Meaning}, and {Actions} {Is} {Related} to {Later}
  {Verb} {Acquisition}.
\newblock {\em Brain Sciences}, 7(12):52.

\bibitem[\protect\astroncite{Oliphant and
  Batali}{1997}]{oliphant_learning_1997}
Oliphant, M. and Batali, J. (1997).
\newblock Learning and the {Emergence} of {Coordinated} {Communication}.
\newblock {\em Center for Research on Language Newsletter}, 11.

\bibitem[\protect\astroncite{Paszke et~al.}{2017}]{paszke_automatic_2017}
Paszke, A., Gross, S., Chintala, S., Chanan, G., Yang, E., DeVito, Z., Lin, Z.,
  Desmaison, A., Antiga, L., and Lerer, A. (2017).
\newblock Automatic differentiation in {PyTorch}.

\bibitem[\protect\astroncite{Peirce}{1998}]{peirce_collected_1998}
Peirce, C.~S. (1998).
\newblock {\em Collected papers of {Charles} {Sanders} {Peirce}}.
\newblock Thoemmes Press, Bristol, England.
\newblock OCLC: ocm39692049.

\bibitem[\protect\astroncite{Rączaszek-Leonardi
  et~al.}{2018}]{raczaszek-leonardi_language_2018}
Rączaszek-Leonardi, J., Nomikou, I., Rohlfing, K.~J., and Deacon, T.~W.
  (2018).
\newblock Language {Development} {From} an {Ecological} {Perspective}:
  {Ecologically} {Valid} {Ways} to {Abstract} {Symbols}.
\newblock {\em Ecological Psychology}, 30(1):39--73.

\bibitem[\protect\astroncite{Skyrms}{2010}]{skyrms_signals:_2010}
Skyrms, B. (2010).
\newblock {\em Signals: evolution, learning, \& information}.
\newblock Oxford University Press, Oxford ; New York.
\newblock OCLC: ocn477256653.

\bibitem[\protect\astroncite{Stern}{1974}]{stern_goal_1974}
Stern, D.~N. (1974).
\newblock The {Goal} and {Structure} of {Mother}-{Infant} {Play}.
\newblock {\em Journal of the American Academy of Child Psychiatry},
  13(3):402--421.

\end{thebibliography}

\clearpage
\appendix        
\section{Dataset}
\begin{figure}[H]
  \centering
\begin{subfigure}{0.19\textwidth}
  \centering
  \includegraphics[width=0.55\linewidth]{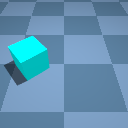}
  \caption{Cyan box}
\end{subfigure}
\begin{subfigure}{0.19\textwidth}
  \centering
  \includegraphics[width=0.55\linewidth]{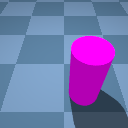}
  \caption{Magenta cylinder}
\end{subfigure}
\begin{subfigure}{0.19\textwidth}
  \centering
  \includegraphics[width=0.55\linewidth]{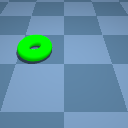}
  \caption{Green torus}
\end{subfigure}
\begin{subfigure}{0.19\textwidth}
  \centering
  \includegraphics[width=0.55\linewidth]{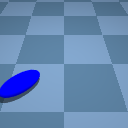}
  \caption{Blue ellipsoid}
\end{subfigure}
\begin{subfigure}{0.19\textwidth}
  \centering
  \includegraphics[width=0.55\linewidth]{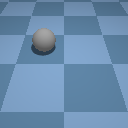}
  \caption{Gray sphere}
\end{subfigure}
\caption{Examples of images from the dataset.}
\label{fig:dataset}
\end{figure}

We conduct our experiments on a dataset consisting of 2500 images of colored three-dimensional objects. Each image has dimensions of 128 x 128 x 3 pixels. The dataset includes images of five shapes (box, sphere, cylinder, torus, ellipsoid) and five colors (blue, cyan, gray, green, magenta). One hundred images generated using POV-Ray ray tracing engine,\footnote{The dataset was generated using code available from \url{https://github.com/benbogin/obverter}.} differing in the position of the object on a surface, are included for each color--shape pair. (An analogous dataset was previously used by \cite{choi_compositional_2018} and \cite{bogin_emergence_2018}.) We choose pairs for the test set by taking one of each figure and color, i.e. the test set is composed of blue boxes, cyan spheres, gray cylinders, green tori and magenta ellipsoids. Example images from the dataset are shown in Figure \ref{fig:dataset}.

\section{Agent architecture}
\paragraph{Vision module}

We pre-train a simple convolutional neural network on the training subset of our datatset to predict colors and shapes. The network is composed of two layers of filters ($20$ and $50$ filters with kernel size $5 x 5$ and stride $1$), each followed by a ReLU (rectified linear unit) activation and max pooling. The output of convolutional layers is then projected into a $25$-dimensional image embedding using a fully-connected layer. During pre-training, the image embedding is passed to two linear classifiers (for color and shape) and the whole vision module is optimized with negative log likelihood as a cost function.
\paragraph{Sender} 

During naming games, the vision module is kept frozen (i.e. it is not updated during training). The sender generates its messages using a single-layer recurrent neural network (RNN) with a hidden state size of 200. The 25-dimensional image embedding for each image is projected to 200 dimensions to initialize the hidden state of the RNN. Let $T$ be a fixed length of the message. Then, at each time-step $t < T$, the output of the RNN is used to parameterize a Gumbel-Softmax distribution (together with a temperature $\tau$ that is a trainable parameter as well). A symbol is sampled from this distribution at each time-step $t$. After reaching $T$, the RNN halts and the generated symbols are concatenated to form a message, which is then passed to the receiver.

\paragraph{Receiver} 

The receiver processes a message symbol-by-symbol using a single-layer recurrent neural network with a hidden state size of 200. After processing the entire sequence, the last output is passed to a two-layer neural network classifier with two softmax outputs for color and shape.

\paragraph{Hyperparameters}
All models are  optimized using Adam \citep{kingma_adam:_2014}. The batch size is always 32. During the pre-traning phase of template transfer, both sender and receiver, as well as the vision classifier, are trained with learning rate $10^{-3}$. During the object naming game, the sender is trained with learning rate $ 10^{-3}$ and receiver with learning rate $10^{-5}$. During experiments with obverter, the receiver is trained with learning rate $10^{-5}$.

\section{Implementations}

Algorithms \ref{algo-baseline}--\ref{algo-obverter} describe the particular implementations of baseline training, template transfer pre-training and obverter used in the experiments.
All experiments are implemented using PyTorch \citep{paszke_automatic_2017} and EGG \citep{kharitonov_egg:_2019}.

\begin{algorithm}[h]
\caption{Baseline training}
\label{algo-baseline}
\begin{algorithmic}[1]
\State Initialize sender $s_\theta$, receiver $r_\psi$,  and training set $\mathcal D$
\For {$x, x_c, x_s \in \mathcal D$}
    \State $m \sim s_\theta(x)$
    \State $\hat{x}_c, \hat{x}_s = r_\psi(m)$
    \State $\mathcal{L}$ = -log\_likelihood($x_c$, $\hat{x}_c$) - log\_likelihood($x_s$, $\hat{x}_s$)
    \State optimize($\mathcal{L}(\theta, \psi)$)
\EndFor
\end{algorithmic}
\end{algorithm}

\begin{algorithm}[h]
\caption{Template transfer}
\label{algo-template transfer}
\begin{algorithmic}[1]
\State Initialize senders $s_{\theta_1}$, $s_{\theta_2}$, $s_\theta,$ receiver $r_\psi$, and training set $\mathcal D$
\For {$x, x_c, x_s \in \mathcal D$}
    \State $m_1 \sim s_{\theta_1}(x)$ \Comment{Color naming game}
    \State $m' \sim$ vocabulary
    \State $\hat{x}_c, \hat{x}_s = r_\psi([m_1, m'])$
    \State $\mathcal{L}_1$ = -log\_likelihood($x_c$, $\hat{x}_c$)
    
    \State $m_2 \sim s_{\theta_2}(x)$ \Comment{Shape naming game}
    \State $m'' \sim$ vocabulary
    \State $\hat{x}_c, \hat{x}_s = r_\psi([m'', m_2])$
    \State $\mathcal{L}_2$ = -log\_likelihood($x_s$, $\hat{x}_s$)
    \State optimize($ (\mathcal{L}_1(\theta_1, \psi) + \mathcal{L}_2(\theta_2, \psi)$))
\EndFor
\For {$x, x_c, x_s \in X$}
    \State $m \sim s_{\theta}(x)$ \Comment{Object naming game}
    \State $\hat{x}_c, \hat{x}_s = r_\psi(m)$
    \State $\mathcal{L}$ = -log\_likelihood($x_c$, $\hat{x}_c$) - log\_likelihood($x_s$, $\hat{x}_s$)
    \State optimize($\mathcal{L}(\theta, \psi)$)
\EndFor
\end{algorithmic}
\end{algorithm}

\begin{algorithm}[h]
\caption{Obverter}
\label{algo-obverter}
\begin{algorithmic}[1]
\State Initialize agents $a_1, a_2$, visual module $v$, training set $\mathcal D$
\State Initialize the set $M$ of all possible messages $m$
\For {$x, x_c, x_s \in \mathcal D$}
    \State $s_\theta, r_\psi \sim \{a_1, a_2\}$ \Comment{Randomly assigning the roles of sender and receiver}
    \State $m = \argmin_{m \in M}$ {evaluate\_message}($s_\theta$, $m$)
    \State $\hat{y}_c, \hat{y}_s = r_\psi(m)$
    \State $\mathcal{L}$ = - log\_likelihood($x_c$, $\hat{x}_c$) - log\_likelihood($x_s$, $\hat{x}_s$)
    \State optimize($\mathcal{L}(\psi)$)
\EndFor
\Procedure{evaluate\_message}{model, $m$}
    \State $x_c, x_s = v(x)$ \Comment{Using visual classifier predictions as a proxy for ground truth labels}
    \State $\hat{x}_c, \hat{x}_s$ = model($m$)
    \State $\mathcal{L}'$ = - log\_likelihood($x_c$, $\hat{x}_c$) - log\_likelihood($x_s$, $\hat{x}_s$)
    \State \Return $\mathcal{L}'$
\EndProcedure
\end{algorithmic}
\end{algorithm}

\section{Loss derivation}
We assume that we are given a 
dataset $\mathcal D =\left\{(x^{(i)}, x_c^{(i)}, x_s^{(i)})\right\}_{i=1}^n$, where entries are i.i.d.,  
$x^{(i)}$ is an RGB image and $x_c^{(i)}$, $y_s^{(i)}$ are (ground truth) labels for  $x^{(i)}$.

We assume that each data-point comes from a distribution
$(X, X_s, X_c, M)$, where $X_s$ and $X_c$ are two labels for image $X$ and $M$ is a latent variable. 

We are interested in minimizing the negative log likelihood of ground truth labels given the image
\begin{equation}
\mathbb{E}_{(x, x_c,x_s)\sim \mathcal D}[-\log p(x_c, x_s|x)].
\end{equation}
%
%(in particular we are considering a discriminative model).
%
%For any probability distribution
%\[
%p(y_1, y_2|x) = \sum_{z}p(y_1, y_2|x,z)p(z|x). 
%\]
We assume that $(X_c,X_s)$ and $X$ are conditionally independent given $M$, i.e. $p(x_s, x_c|x,m) = p(x_s, x_c|m)$. Consequently, 

\begin{equation}
\log p(x_c, x_s|x) = \log \sum_{m}p(x_c, x_s|m)p(m|x)
\ge \mathbb{E}_{m\sim p(\cdot|x)}[\log p(x_c, x_s|m)] 
\end{equation}

where the last inequality follows from Jensen's inequality.
We will be optimizing the lower bound; hence, our surrogate loss function is 
\begin{equation}
\mathbb{E}_{(x, x_c,x_s)\sim \mathcal D}\mathbb{E}_{m\sim p(\cdot|x)}[-\log p(x_c, x_s|m)].
\end{equation}

During template transfer pre-training (color naming game and shape naming game) we additionally assume that $X_c$ is conditionally independent from $X_s$ given $X$, i.e. 
\begin{equation}
p_{\theta_1, \theta_2, \psi}(x_c,x_s|x)=p_{\theta_1, \psi}(x_s|x)p_{\theta_2, \psi}(x_c|x).
\end{equation}
This implies that the surrogate loss is equal to
\begin{equation}
\mathbb{E}_{(x_c,x)\sim \mathcal D}[-\log p_{\theta_1, \psi}(x_c|x)]
+\mathbb{E}_{(x_s,x)\sim \mathcal D}[-\log p_{\theta_2, \psi}(x_s|x)]. 
\end{equation}

Furthermore, we assume that
\begin{equation}
p_{\theta_1, \psi} (x_c|x)=\sum_{m}s_{\theta_1}(m|x)r_\psi(x_c|m), \qquad 
p_{\theta_2, \psi}(x_s|x)=\sum_{m}s_{\theta_2}(m|x)r_\psi(x_s|m),
\end{equation}
where both marginal distributions share the same $r_\psi$. 

During the object naming game (the second phase of template transfer), we have
\begin{equation}
p_{\theta_3,\psi}(x_c, x_s|x)=\sum_{m}s_{\theta_3}(m|x)r_\psi(x_c, x_s|m), 
\end{equation}
where $r_\psi$ is the same as in the color naming and shape naming games.

\end{document}